\documentclass[12pt]{article}
\usepackage{geometry}
\usepackage[parfill]{parskip} % Activate to begin paragraphs with an empty line rather than an indent
\usepackage{graphicx} % Use pdf, png, jpg, or eps with pdflatex; use eps in DVI mode
\usepackage{amssymb,amsmath,amsthm}
\usepackage{commath}
\usepackage{longtable}
\usepackage{tabularx}
\usepackage[hyphens]{url}
\usepackage[dvipsnames]{xcolor}
\usepackage[unicode=true,colorlinks=true,urlcolor=CadetBlue,citecolor=black,linkcolor=black]{hyperref}
\usepackage{booktabs}
\usepackage{pgfplotstable} % http://mirrors.ctan.org/graphics/pgf/contrib/pgfplots/doc/pgfplotstable.pdf
\PassOptionsToPackage{hyphens}{url} % url is loaded by hyperref
\usepackage{authblk}
\usepackage{subcaption}
\usepackage{tikz}
\usepackage[font=small,labelfont=bf]{caption}
\usepackage{gensymb}
\usepackage{cleveref}
\usepackage[table]{xcolor}
\usepackage{bibunits}
\usepackage{comment}
\definecolor{salorange}{HTML}{F28C28}
\definecolor{salorangeDark}{HTML}{ec6e07}

\newcommand{\tablestyle}{\small\renewcommand{\arraystretch}{1}\setlength{\aboverulesep}{0pt}\setlength{\belowrulesep}{0pt}}
\newcommand{\tabtop}{\specialrule{1pt}{0pt}{0pt}}
\newcommand{\tabhead}{\specialrule{0.6pt}{0pt}{0pt}}
\newcommand{\tabsep}{\specialrule{0.4pt}{0pt}{0pt}}
\newcommand{\tabbottom}{\specialrule{1pt}{0pt}{0pt}}
\newcommand{\grp}[1]{\textbf{#1}}

\excludecomment{draftnotes}

\usepackage{newtxtext} %loads helv for ss, txtt for tt
\usepackage[bigdelims]{newtxmath}
\usepackage[T1]{fontenc}
\usepackage{textcomp}
\makeatletter
 \renewcommand\section{\@startsection {section}{1}{\z@}%
     {-2.5ex \@plus -1ex \@minus -.2ex}%
     {1.3ex \@plus.2ex}%
    {\Large\bfseries}}
    
\usepackage{xspace}

\newcommand{\species}[3]{%
  \newcommand{#1}{\gdef#1{\textit{#3}\xspace}\textit{#2}\xspace}}

\species{\eastsalamander}{Salamandra infraimmaculata}{S.~infraimmaculata}
\species{\eurosalamander}{Salamandra salamandra}{S.~salamandra}
\species{\latonia}{Latonia nigriventer}{L.~nigriventer}
\species{\snowleopard}{Panthera uncia}{P.~uncia}
\species{\japsalamander}{Andrias japonicus}{A.~japonicus}

\newcommand{\beginsupplement}{%
      	\setcounter{table}{0}
        \renewcommand{\thetable}{S\arabic{table}}%
        \setcounter{figure}{0}
        \renewcommand{\thefigure}{S\arabic{figure}}%
}

\usepackage[round,colon,authoryear]{natbib}

\title{Accurate automated photo-identification of Near Eastern and European fire salamanders}

\author[1]{Maayan Yesharim}
\author[2]{Hava Goldstein}
\author[3]{Gabi Kolodny}
\author[3]{Assaf Ben-Levy}
\author[3]{Tamar Kis-Papo}
\author[3]{Dor Gavish}
\author[3]{Rotem Vidan}
\author[3]{Nirit Lavie-Alon}
\author[2,3]{Olga Rybak}
\author[1,4,*]{Yoav Ram}

\affil[1]{School of Zoology, Faculty of Life Sciences, Tel Aviv University, Tel Aviv, Israel}
\affil[2]{Israel Nature and Parks Authority, Israel} 
\affil[3]{The Society for the Protection of Nature in Israel, Tel Aviv, Israel}
\affil[4]{Safra Center for Bioinformatics, Tel Aviv University, Tel Aviv, Israel}
\affil[*]{Corresponding author: yoavram@tauex.tau.ac.il}

\begin{document}
\maketitle

% Keywords: re-identification; local-feature matching; capture--recapture; amphibian; conservation

%%%%%%%%%%%%%%%%%%%%%%%%%%%%%%
\begin{abstract}
\linespread{1.3}\selectfont
Photo-identification underpins non-invasive capture--recapture monitoring of amphibians, but manual identification becomes slow and error-prone as reference libraries grow.
Focusing on the Near Eastern fire salamander, \eastsalamander, we developed a pipeline based on zero-shot local-feature matching (SuperPoint and LightGlue) without species-specific training and evaluated it on $2{,}440$ field photographs from Haifa and Tel Dan in northern Israel.
We isolated the yellow spot pattern, scored image similarity by the number of high-confidence keypoint matches, which reduced a confound whereby many-spotted individuals accumulate spurious matches, and assigned identities across the database by two-tier clustering.
Pipeline-assisted expert review corrected misidentifications, reducing the number of identities by almost a quarter.
On held-out individuals, the pipeline ranked the correct individual first for $95.2\%$ of queries and recovered identities with an ARI of $0.958$.
Without modification, it reached $95.1\%$ and $91.7\%$ top-1 accuracy on two datasets of the European fire salamander, \eurosalamander, and recognised $93.7\%$ of recaptures in one of them, compared with $38$--$77\%$ for five established programs and $93.9\%$ for visual inspection.
Zero-shot local-feature matching thus offers amphibian monitoring a re-identification tool that needs no training data, transfers between related species, scales to databases of over a thousand individuals, and can audit existing databases before capture--recapture analysis.
\end{abstract}

\newpage
%%%%%%%%%%%%%%%%%%%%%%%%%%%%%%
\section*{Introduction}

% ---- 1. The species
Amphibians are the most threatened class of vertebrates \citep{stuart2004status}. The Near Eastern fire salamander, \eastsalamander, reaches the southernmost limit of its genus in Israel \citep{goedbloed2017evaluation}, where it is considered endangered \citep{meiri2019checklist}, although it is listed globally as Least Concern, previously Near Threatened \citep{iucn2022salamandra}. Adults are terrestrial and breed in ephemeral pools and permanent springs \citep{segev2010population}, where introduced mosquitofish harm their larvae \citep{segev2009deleterious}, and several populations persist in fragmented and urban habitats \citep{bardavid2007long, darel2025effective}. Each individual carries a unique pattern of discrete yellow spots on its back, which allows it to be recognized from photographs without physical marking \citep{darel2025effective}. In the congeneric \eurosalamander, this pattern is widely used for identification, but some adults show minor changes, such as the shrinking or splitting of a spot, possibly linked to growth \citep{balogova2016changes}.

% ---- 2. Earlier work on the species in Israel
Dorsal spot patterns have been used to recognize individuals of this species for decades. On Mount Carmel, an isolated population was followed by its dorsal patterns for 18 years \citep{warburg1994population}, and individually recognized salamanders were followed for up to 21 years, over which most growth occurred in the first 9--10 years of life \citep{warburg2008growth}. Capture--recapture studies also identified individuals by eye from their dorsal patterns \citep{segev2010population}, and movement studies revealed long-distance dispersal with implications for habitat fragmentation \citep{bardavid2007long}. In a first test of automated identification in this species, the AmphIdent software missed 47\% of the recaptures that a trained observer had found in the photographs of \citet{segev2010population}, which were taken at varying angles, but matched recaptures reliably in photographs taken perpendicular to the back under a standardized protocol at Tel Dan \citep{goedbloed2017evaluation}. Most recently, a multi-year citizen-science program in two urban wadis of Haifa recorded 212 individuals over 89 surveys and estimated population size, survival and growth, revealing a prolonged period of vulnerability before individuals reach full size and smaller populations where breeding ponds contain introduced fish \citep{darel2025effective}. In that program, recaptures were recognized by a single observer comparing photographs by eye, and the authors noted that automated identification tools would make this step robust and repeatable \citep{darel2025effective}.

% ---- 3a. Photo-identification, manual matching and classical software
Photo-identification is now widely used for non-invasive capture--recapture studies \citep{matthe2017comparison}, and for many amphibians it is the basis of long-term monitoring of survival, movement and population size \citep{bendik2013computer, drechsler2015genetic, schulte2024performance, yesharim2026latonia}. However, manual matching becomes time-consuming and error-prone as the reference library grows over multiple surveys \citep{bolger2012computer}. Most such errors are splitting errors, in which a recaptured individual is registered as new, and they bias capture--recapture estimates of population size upward \citep{stevick2001errors, johansson2020identification, kodi2024ghostbusting}; in a critically endangered skink, supplementing human matching with software revealed such errors and lowered abundance estimates \citep{jordan2025supplementing}. Computer-assisted systems based on hand-crafted features were developed to reduce this burden, including for salamanders \citep{gamble2008multi, bolger2012computer, crall2013hotspotter, drechsler2015genetic}. In a large capture--recapture study of newts, genetic fingerprinting confirmed that one such program, AmphIdent, identified individuals highly efficiently \citep{drechsler2015genetic}, but in adult European fire salamanders photographed in the field, five such programs recognised only 38--77\% of recaptures \citep{schulte2024performance}.

% ---- 3b. Deep-learning re-identification
Automated re-identification is increasingly used in wildlife monitoring \citep{delorm2023optimizing}, and recent deep-learning approaches broadly fall into two families: global-feature embedding models, which represent each image by a single vector, such as MegaDescriptor and MiewID \citep{cermak2024wildlifedatasets, adam2025wildlifereid, otarashvili2024miewid}, and deep local-feature matching pipelines, which detect learned keypoints and match them between images \citep{detone2018superpoint, zhao2023aliked, lindenberger2023lightglue} and have been applied to individual re-identification on their own \citep{yesharim2026latonia} or combined with global embeddings \citep{nepovinnykh2024species, cermak2025wildfusion}. In the AnimalCLEF 2025 challenge, which included European fire salamanders, most top-ranked systems included local-feature matching \citep{adam2025overview}; the 2026 edition changed the task to clustering images into individuals, evaluated by the adjusted Rand index \citep{adam2026overview}. In amphibians, deep networks trained on the target species have identified individual Japanese giant salamanders, \japsalamander, and yellow-spotted mountain newts \citep{takaya2023individual, rahmdel2026deep}, but such classifiers recognise only the individuals in their training images, whereas wild populations continually gain new individuals.

% ---- 4. The Hula painted frog study
In our previous study, zero-shot local-feature matching, a pretrained keypoint detector and matcher with candidates ranked by the number of matches, identified the critically endangered Hula painted frog (\textit{Latonia nigriventer}; \citealt{iucn2021latonia}) from its ventral pattern with $99.8\%$ top-1 accuracy ($1{,}232$ images of $191$ individuals), without species-specific training, whereas a fine-tuned global-embedding model was far less accurate \citep{yesharim2026latonia}. Two questions remain open: whether the approach transfers to an amphibian with a different body plan and pattern, which may require adapting the preprocessing; and how to assign identities automatically across a whole database and evaluate the result, which the frog study did not address in depth.

% ---- 5. This study
Here we apply zero-shot local-feature matching to a photo-identification database of $2{,}440$ images of \eastsalamander from four sites in northern Israel, taken by hand at close range, with identities assigned by field researchers through visual comparison against a growing reference library. We (1) use an early version of the pipeline to audit these labels, which reduced the number of identities from $1{,}431$ to $1{,}094$; and, on the verified labels, (2) adapt the preprocessing to the target species and score image pairwise similarity by keypoint matching confidence, which reduces a confound between the number of spots and the similarity score; (3) add a two-tier clustering that assigns identities across the database, with all choices tuned on a validation set and evaluated on held-out individuals; and (4) apply the pipeline, unchanged, to two datasets of European fire salamanders and compare it with five established photo-identification programs.

% \newpage
\section*{Methods}

%%%% DATASETS
\subsection*{Datasets: Near Eastern fire salamander}

This study draws on several datasets of labelled images of the Near Eastern fire salamander (\eastsalamander), collected during nocturnal field surveys at three sites in the Haifa area and one at Tel Dan Nature Reserve, northern Israel. 
The Haifa datasets are labelled AH, from the Wadi Ahuza site, and HFA, from two sites, Wadi Lotem and Haifa Educational Zoo, as individual movement between the two sites has been observed. These datasets were collected during surveys across four winters between December 2021 and December 2025 by a long-term citizen science monitoring project \citep{darel2025effective}.
The Tel Dan datasets (NTD\,2021--22 and NTD\,2022--23) were collected during surveys conducted over two bounded winter seasons by Israel Nature and Parks Authority (INPA).

Photos were taken using flash phone cameras at a distance of 40--60 cm with the animal in dorsal view to capture yellow spot patterns and saved as colour JPG images (Figure~\ref{fig:Fig1}a,b).
Some photos contained partial occlusion from mud or vegetation.
In total, the datasets include $2{,}440$ images of $1{,}431$ putative individuals, with identity labels assigned by field researchers through visual comparison of each new image against a reference library of previously identified individuals.
Following an identity audit (see Results), labels were reassigned to $1{,}094$ verified individuals (Table~\ref{tab:dataset}).

The Haifa 2020-2022 surveys were conducted under the INPA permits 42648 and 42905.
The Tel Dan surveys were conducted by INPA staff and required no permit.

\paragraph{Dataset split.}
The combined Near Eastern fire salamander dataset (merge of HFA, AH, NTD) was used for pipeline development, hyper-parameter tuning, and evaluation.
To prevent leakage between hyper-parameter tuning and evaluation, the combined dataset was partitioned at the individual level into a validation set (984 individuals, 2,182 images) and a held-out test set (110 individuals, 258 images).
The test set was drawn only from the Haifa datasets (AH and HFA): when the split was defined, the Tel Dan labels were suspected to contain errors, and the identity audit confirmed that nearly all label errors were there (see Results: \textit{Identity audit}).
As a result, the test-set labels are nearly independent of the pipeline-assisted audit, which changed only a single Haifa label.
All scoring and threshold decisions were made on the validation set; the test set was used only once, after all pipeline decisions were fixed, to confirm generalisation.
Final performance metrics are reported on the full dataset (1,094 individuals, 2,440 images), evaluated with the optimized pipeline. This partition is consistent with the split used for head-orientation classifier training (see \textit{Head-orientation classifier training}), ensuring no individual appears in both the parameter-tuning set and the held-out test set.

\subsection*{Datasets: European fire salamander}

To evaluate the generalisation of our pipeline beyond the study populations and species, we applied it without modification to two independent datasets of the European fire salamander (\eurosalamander).

\begin{figure}
    \centering
    \begin{subfigure}{0.49\linewidth}
        \begin{tikzpicture}
            \node[anchor=south west, inner sep=0] (image) at (0,0)
                {\includegraphics[height=3.7cm, keepaspectratio]{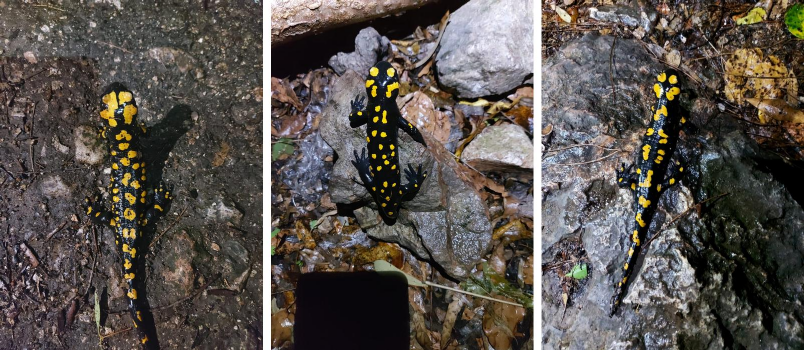}};
            \node[anchor=south west, fill=white, fill opacity=0.6, 
                  text opacity=1, font=\small] at (image.south west) {(a)};
        \end{tikzpicture}
    \end{subfigure}
    \hfill
    \begin{subfigure}{0.49\linewidth}
        \begin{tikzpicture}
            \node[anchor=south west, inner sep=0] (image) at (0,0)
                {\includegraphics[height=3.7cm, keepaspectratio]{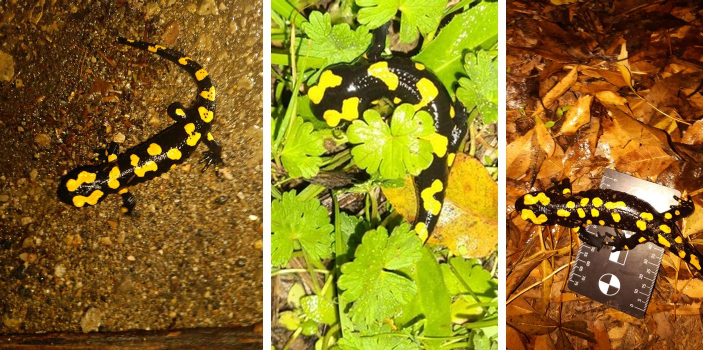}};
            \node[anchor=south west, fill=white, fill opacity=0.6, 
                  text opacity=1, font=\small] at (image.south west) {(b)};
        \end{tikzpicture}
    \end{subfigure}

    % \vspace{0.5em}

    \begin{subfigure}{0.49\linewidth}
        \begin{tikzpicture}
            \node[anchor=south west, inner sep=0] (image) at (0,0)
                {\includegraphics[width=\linewidth, height=3.8cm, keepaspectratio]{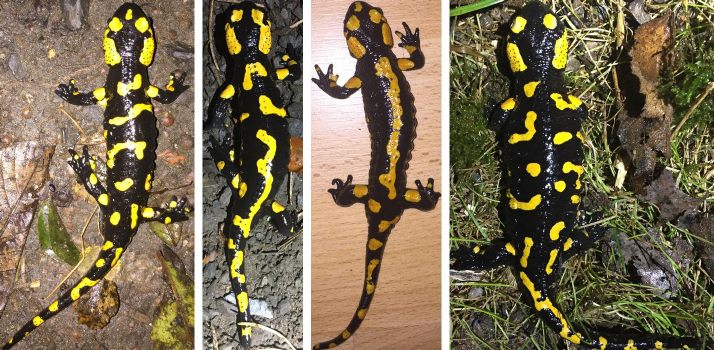}};
            \node[anchor=south west, fill=white, fill opacity=0.6, 
                  text opacity=1, font=\small] at (image.south west) {(c)};
        \end{tikzpicture}
    \end{subfigure}
    \hfill
    \begin{subfigure}{0.49\linewidth}
        \begin{tikzpicture}
            \node[anchor=south west, inner sep=0] (image) at (0,0)
                {\includegraphics[width=\linewidth, height=3.8cm, keepaspectratio]{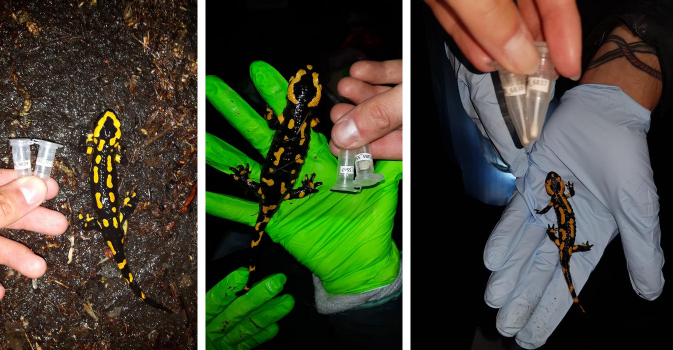}};
            \node[anchor=south west, fill=white, fill opacity=0.6, 
                  text opacity=1, font=\small] at (image.south west) {(d)};
        \end{tikzpicture}
    \end{subfigure}

    \caption{\textbf{Fire salamander dorsal-view photos.}
    Each panel shows 3-4 individuals from one of the datasets of the two species:
    Near Eastern fire salamander (\eastsalamander) from Haifa (a) and Tel Dan Nature Reserve (b). European fire salamander (\eurosalamander) from the Czech Republic (c) and Germany (d).
    Note the visual difference between species: \eastsalamander displays discrete yellow spots, while \eurosalamander displays variable markings ranging from discrete spots to elongated stripes.}
    \label{fig:Fig1}
\end{figure}

\paragraph{Czech dataset.}
Images of European fire salamander (\eurosalamander) from the Czech Republic were released as part of the AnimalCLEF 2025 benchmark \citep{adam2025overview}.
Images were collected during nocturnal field surveys under rainy conditions between 2017 and 2024 (per date field in the dataset metadata), with individuals photographed and immediately released without marking (Figure~\ref{fig:Fig1}c).
We only used dorsal view images, with 784 images of 498 individuals. 

\paragraph{German dataset.}
A second dataset of European fire salamander \textit{S.~salamandra} was collected between 2018 and 2020 in the Kruppwald forest, Essen, Germany \citep{schulte2024performance} (Figure~\ref{fig:Fig1}d). Images were taken opportunistically in the field under non-standardised conditions with varying backgrounds, and were provided with the head oriented upward. Identity labels were also provided by the authors of the original study. In their count, one image was duplicated and three recaptures were counted twice; after correcting these, the dataset comprises $376$ images of $313$ individuals with $63$ recaptures.

\begin{table}[h]
\centering
\tablestyle
\begin{tabular}{lrrrrr}
\tabtop
\textbf{Dataset} & \textbf{Site} & \textbf{IDs} & \textbf{Images} & \textbf{Singletons} & \textbf{Survey nights}\\
\tabhead
\multicolumn{6}{l}{\textbf{Near Eastern fire salamander}} \\
HFA & Wadi Lotem \& Haifa Educational Zoo & 37  & 134   & 11 & 17  \\
AH & Haifa: Wadi Ahuza & 182 & 376   & 97 & 25 \\
NTD\,2021--22 & Tel Dan Nature Reserve & 322 & 590 & 179 & 8 \\
NTD\,2022--23 & Tel Dan Nature Reserve& 553 & 1{,}340 & 298 & 11 \\
% \tabsep
\textbf{Total} & Israel & 1,094 & 2,440 & 585 & 61 \\
\tabsep
\multicolumn{6}{l}{\textbf{European fire salamander}} \\
Adam et al. & Czechia & 498 & 784 & 338 & 189 \\ 
Schulte et al. & Kruppwald forest, Essen, Germany & 313 & 376 & 267 & not reported \\ 
\tabbottom
\end{tabular}
\caption{\textbf{Fire salamander datasets used in this study.} \label{tab:dataset}
Singletons: number of individuals with a single image in the database, excluded as queries in accuracy evaluation.
Number of individuals in HFA, AH, and NTD is after identity audit.
}
\end{table}

\subsection*{Pipeline overview}
The re-identification pipeline segments each image, resolves head orientation, extracts the yellow spot pattern, matches keypoints between images to score pairwise image similarity, and uses the similarities to cluster images to individual identities (all described below). 
Head orientation is resolved by a CNN classifier trained on heuristic orientation labels with manual corrections (see \textit{Head-orientation classifier training}). This classifier did not yet exist when the identity audit was run (see Results: \textit{Identity audit}): the audit used the \textit{experimental pipeline}, a preliminary version that resolved orientation with a geometric heuristic instead, and used untuned defaults for keypoint extraction and match scoring. We refer to the experimental pipeline below only where it differs from the \textit{optimized pipeline}, the final version described here, in ways that matter for the results that follow.

\subsection*{Image preprocessing}
\paragraph{Segmentation.} 
All images were segmented using SAM~3 \citep{carion2025sam3segmentconcepts} with the text prompt \emph{``salamander''}. The resulting mask was used to fill background pixels with black, and the image was cropped to the minimal bounding box around the mask.

\paragraph{Head orientation.}
Consistent head orientation is required for reliable keypoint matching.
We aligned all images so the head faces the left side, using two steps. First, PCA is applied to the SAM~3 mask to align the body's long axis to the horizontal axis, placing the head on either the left or right side of the image.
Second, the left--right ambiguity is resolved by the custom CNN classifier (see below).
Then, images classified as right-facing are automatically rotated $180^\circ$.

The experimental pipeline, used to run the identity audit before the CNN classifier was developed, resolved this ambiguity with a geometric heuristic instead: the rotated mask was divided into left and right halves along the body axis, and the longest unbroken vertical run of mask pixels in the single central column of each half was measured. The side with the taller run was assumed to be the head, because a head cross-section is taller than the tapering tail. Images were then rotated $180^\circ$ if the head was determined to be on the right side (ties were left unchanged).

\paragraph{Head-orientation: classifier training.}
We trained a convolutional neural network (CNN) classifier to predict head side, left vs.\ right, from the PCA-aligned, masked image.
We used ResNet-18 \citep{he2016residual} pretrained on ImageNet-1K \citep{deng2009imagenet} as the backbone with a linear classifier head.
Training labels were the orientations assigned by the geometric heuristic, corrected manually in two passes rather than through a single systematic audit: during pipeline development, every query or top-ranked gallery image implicated in a top-1 identification error was inspected and rotated $180^\circ$ if mis-oriented; additional errors were corrected opportunistically when noticed during the identity audit (see Results), but this second pass was not exhaustive.
Each image, after correction, was treated as a head-left example (label 1); a $180^\circ$ rotation of the same image served as the head-right example (label 0), doubling the effective training set. The validation set (see \textit{Dataset split}) was further divided at the individual level into a training subset (875 individuals, 1,930 images) and a model-selection subset (109 individuals, 252 images); the held-out test set (110 individuals, 258 images) was used only to report final accuracy.

Input images were padded to a square aspect ratio, preserving the salamander's elongated proportions, before resizing to $224 \times 224$\,px per ResNet-18 specifications.
During training, images were augmented with random rotation in the range of $\pm10^\circ$, colour jitter (brightness and contrast scaled by a factor in $[0.7, 1.3]$), and Gaussian blur (kernel standard deviation $\sigma \in [0.1, 1.0]$), each applied to every training image; vertical flip was additionally applied with probability $0.5$. Horizontal flip was not used, as it would invert the head-direction label. The model was trained with binary cross-entropy loss and the AdamW optimiser (initial learning rate $1\times10^{-4}$), with learning rate halved after three epochs without improvement in model-selection loss, for 25 epochs.
The checkpoint with the lowest model-selection loss was kept and used for inference, reaching 99.6\% accuracy on the held-out test set (514/516 correct; each image in both orientations).
The checkpoint is available on Hugging Face at \url{https://huggingface.co/yoavram-lab/salamandra-head-orientation}.

\paragraph{Yellow spot extraction.}
After orientation, the salamander's yellow spot pattern was isolated using LAB colour-space thresholding. The image was converted from RGB to LAB, and the $b^*$ channel (blue--yellow axis) was thresholded to retain yellow pixels.
The threshold was determined per image using Otsu's method \citep{otsu1979threshold}, applied only to $b*$ values sampled from an eroded version of the animal mask to exclude the body outline from threshold estimation.
The resulting binary mask was cleaned with morphological opening to remove speckle noise, followed by closing and hole-filling to consolidate each spot into a solid region.

\begin{figure}[htbp]
    \centering
    % Row 1: three images, same height ratio, forced to fit line width
    \resizebox{\linewidth}{!}{%
        \begin{tikzpicture}
            \node[anchor=north west, inner sep=0] (image) at (0,0)
                {\includegraphics[height=3.5cm,]{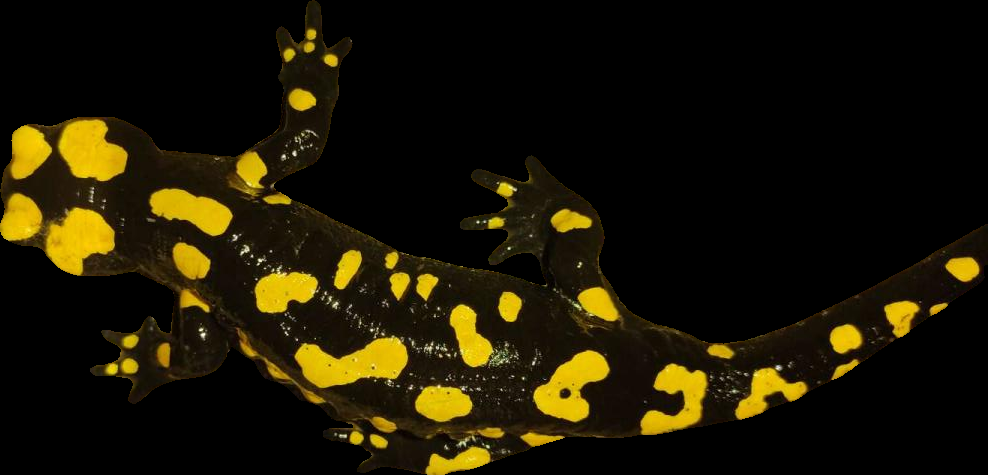}}; %ID 349
            \node[anchor=north west, fill=white, fill opacity=0.6, 
                  text opacity=1, font=\small] at (image.north west) {(a)};
        \end{tikzpicture}
        \hfill
        \begin{tikzpicture}
            \node[anchor=north west, inner sep=0] (image) at (0,0)
                {\includegraphics[height=3.5cm]{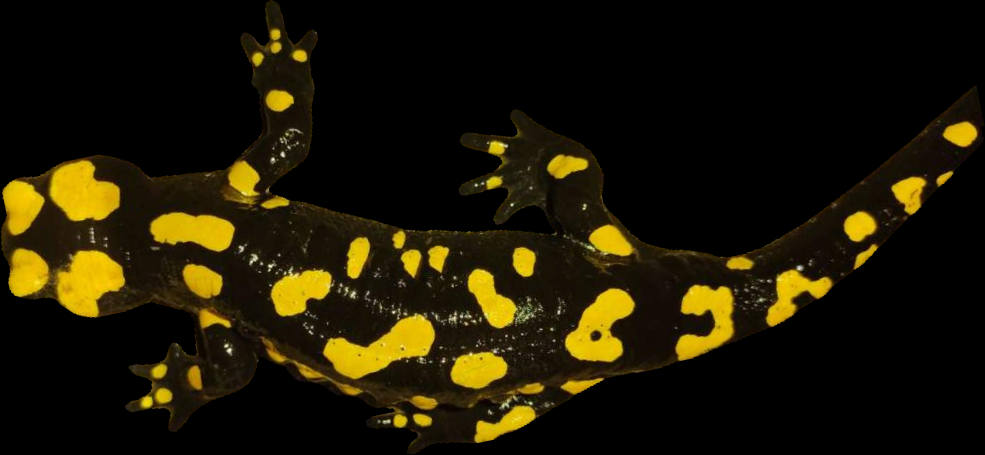}};
            \node[anchor=north west, fill=white, fill opacity=0.6, 
                  text opacity=1, font=\small] at (image.north west) {(b)};
        \end{tikzpicture}
        \hfill
        \begin{tikzpicture}
            \node[anchor=north west, inner sep=0] (image) at (0,0)
                {\includegraphics[height=3.5cm]{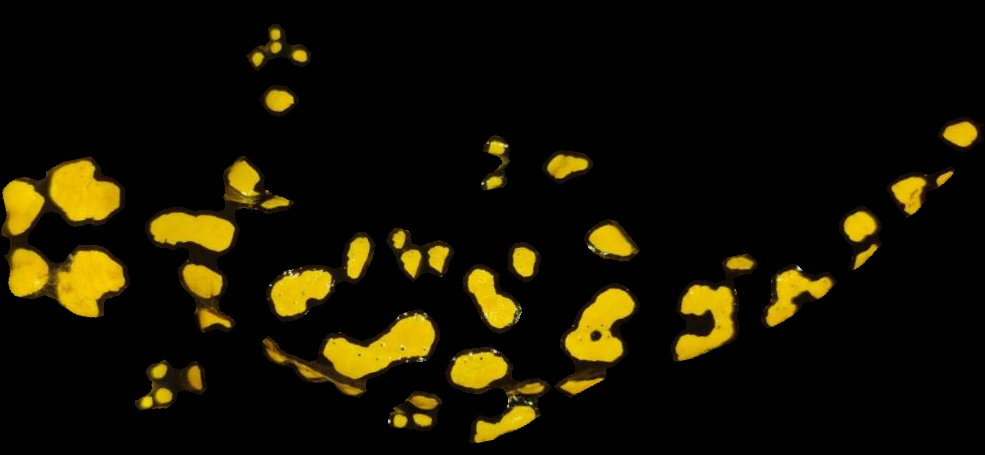}};
            \node[anchor=north west, fill=white, fill opacity=0.6, 
                  text opacity=1, font=\small] at (image.north west) {(c)};
        \end{tikzpicture}%
    }
    % Row 2: three images, same height ratio, forced to fit line width
    \resizebox{\linewidth}{!}{%
        \begin{tikzpicture}
            \node[anchor=north west, inner sep=0] (image) at (0,0)
                {\includegraphics[height=3.5cm,angle=180]{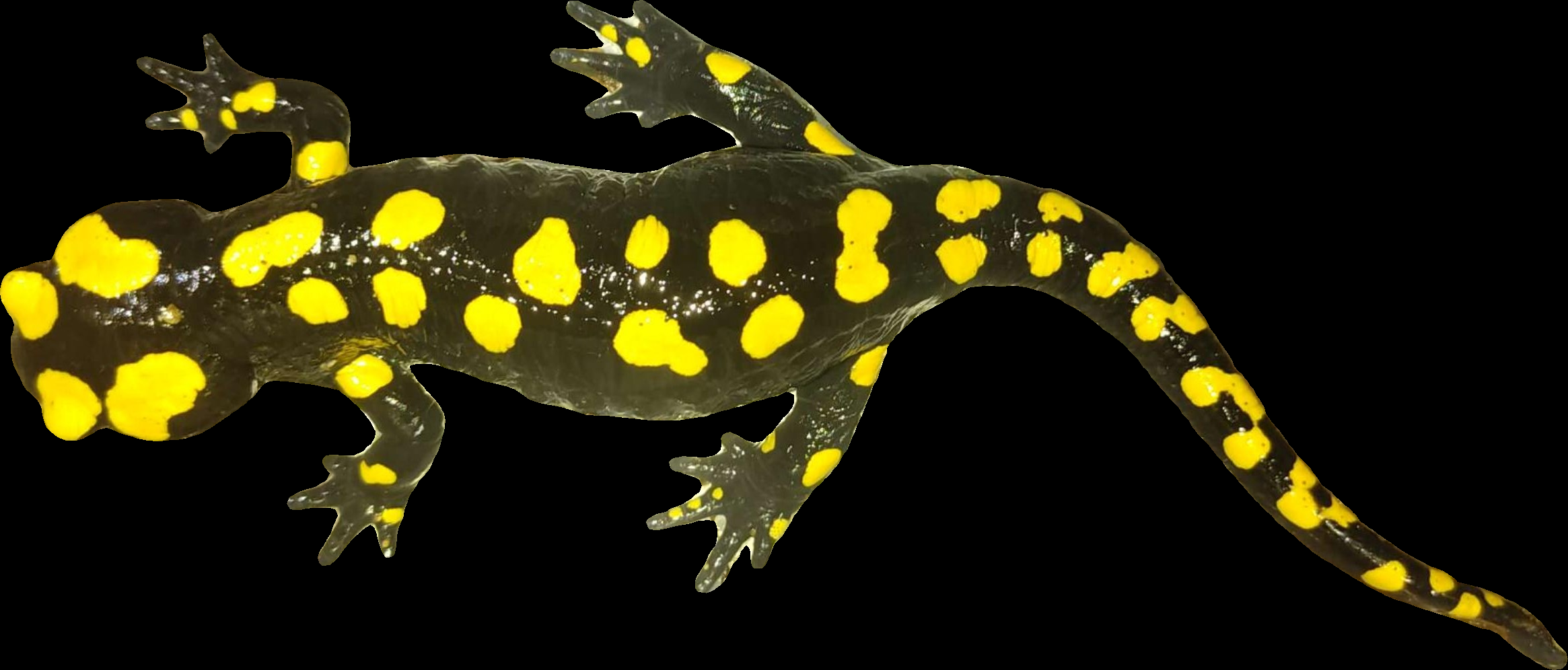}};
            \node[anchor=north west, fill=white, fill opacity=0.6, 
                  text opacity=1, font=\small] at (image.north west) {(d)};
        \end{tikzpicture}
        \hfill
        \begin{tikzpicture}
            \node[anchor=north west, inner sep=0] (image) at (0,0)
                {\includegraphics[height=3.5cm]{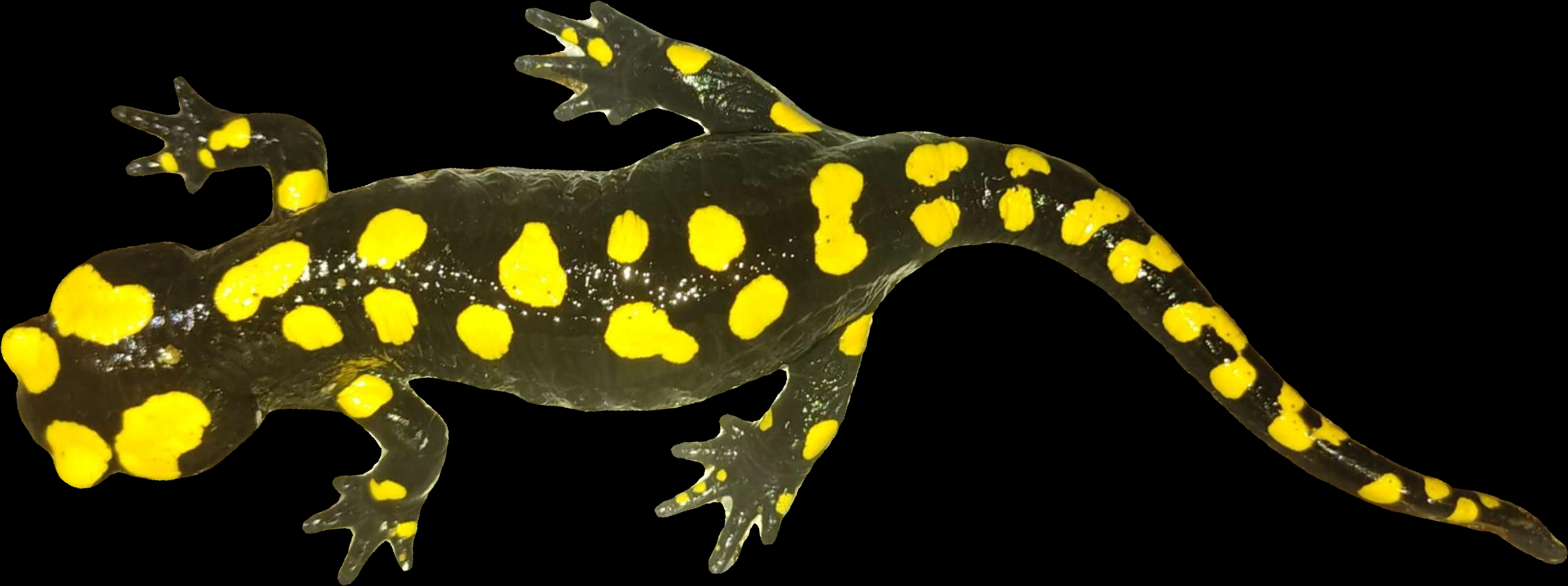}};
            \node[anchor=north west, fill=white, fill opacity=0.6, 
                  text opacity=1, font=\small] at (image.north west) {(e)};
        \end{tikzpicture}
        \hfill
        \begin{tikzpicture}
            \node[anchor=north west, inner sep=0] (image) at (0,0)
                {\includegraphics[height=3.5cm]{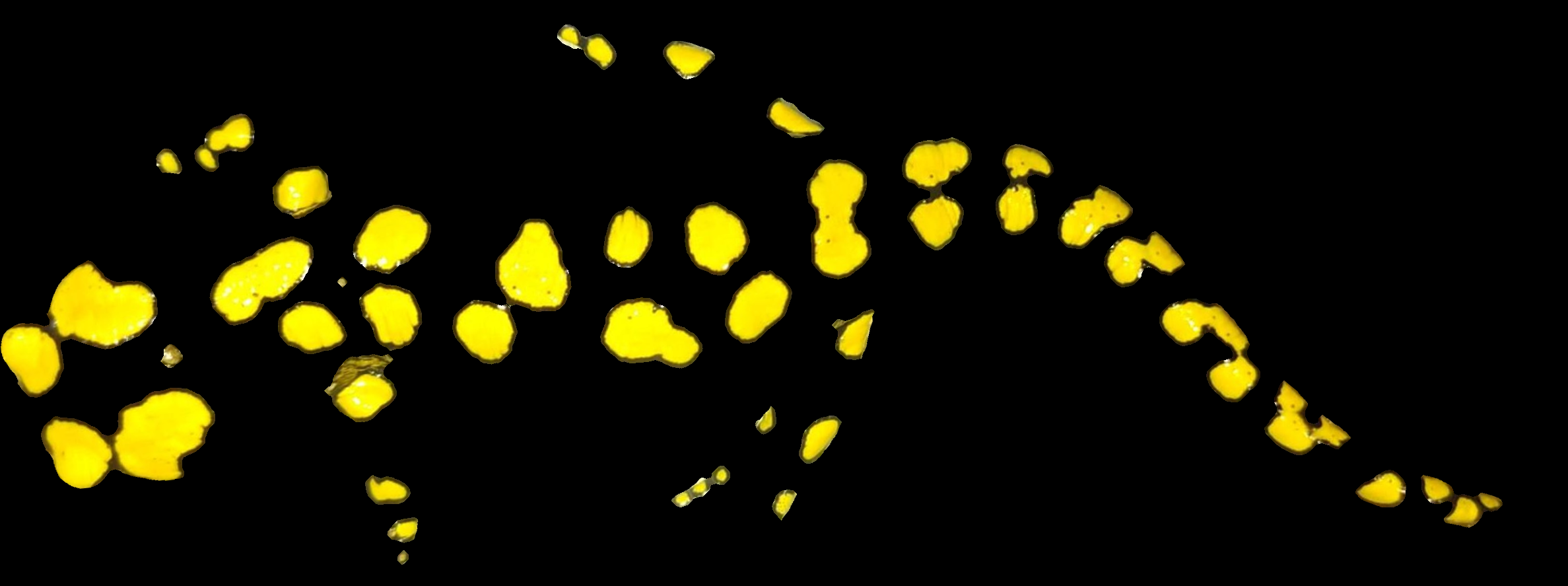}};
            \node[anchor=north west, fill=white, fill opacity=0.6, 
                  text opacity=1, font=\small] at (image.north west) {(f)};
        \end{tikzpicture}%
    }
    \caption{\textbf{Image preprocessing applied to two examples.} 
    Each row shows one individual. \textbf{(a,~d)} Segmentation of body. \textbf{(b,~e)} Head orientation: in (a)$\rightarrow$(b) only a subtle axis alignment is applied; in (d)$\rightarrow$(e) the classifier identifies head-on-right-side and the image is rotated $180^\circ$. \textbf{(c,~f)} Yellow spot extraction. (a,b,c) is individual 349 image 416; (d,e,f) is individual 307 image 228.} %paths: data/photos_mask/w22_307/307NTD228_2022-01-23.png, data/photos_mask/w22_349/349NTD416_2022-02-04.png 
    \label{fig:preprocessing}
\end{figure}

%%%%% Local-feature matching
\subsection*{Local-feature matching}
We use local-feature matching, which outperformed global embedding models in our recent study of \latonia, the Hula painted frog \citep{yesharim2026latonia}. 
Local-feature matching represents each image by a set of keypoints (points of interest) and their descriptors (vectors summarising the image appearance around each keypoint), and then matches keypoints between pairs of images. 
It thus has two steps: descriptor extraction, applied to each image, and descriptor matching, applied to each pair of images.
We applied the extractor and matcher without additional training or fine-tuning on species-specific or any other wildlife data.
Keypoint extraction and matching ran on a single NVIDIA RTX A4000 GPU (16\,GB).

\paragraph{Descriptor extraction.}
For each image we detect keypoints and compute a descriptor for each one. 
We use \emph{SuperPoint}, a fully convolutional neural network (1.3M parameters) that jointly detects keypoints and computes fixed-length keypoint descriptors in a single forward pass, outputting a set of keypoints ($\{x,y\}$ positions in the input image) and their descriptor vectors \citep{detone2018superpoint}. 
SuperPoint uses a shared encoder (VGG-style CNN) that feeds into two decoder heads: one for keypoint detection and one for descriptor extraction, producing 256-dimensional descriptors. 
SuperPoint is trained in three stages: a base detector is first trained on a synthetic dataset of simple geometric shapes with unambiguous corner locations; this detector is then applied to many randomly homography-warped copies of unlabelled real images (MS-COCO), and the detector response maps are warped back to the original image coordinates and averaged to produce pseudo-ground-truth keypoints (\emph{homographic adaptation}); finally, the detector and descriptor heads are trained jointly on image pairs related by known random homographies, using the pseudo-ground-truth keypoints for detection and the homography-induced correspondences for the descriptor loss \citep{detone2018superpoint}.
We used up to $k=400$ keypoints per image and SuperPoint's default detection threshold ($0.0005$); the experimental pipeline used $k=800$.

\paragraph{Descriptor matching.}
The extracted keypoints are matched across image pairs using \emph{LightGlue}, a transformer-based local-feature matcher (11.9M parameters) \citep{lindenberger2023lightglue}. 
We used the publicly released SuperPoint-specific LightGlue weights, which were pre-trained on synthetic homographies generated from the Oxford--Paris 1M distractors dataset and then fine-tuned on MegaDepth, a dataset of crowd-sourced landmark photographs; no species-specific or even wildlife training was performed: the model was trained on landmarks. 
LightGlue jointly processes the two keypoint sets with interleaved self- and cross-attention layers, and then predicts a soft partial match-confidence matrix from pairwise descriptor similarity and per-keypoint matchability. 
Keypoint pairs that are mutual best matches with a confidence score above a threshold are retained as matches, and their confidence scores serve as the basis for the downstream pairwise image similarity score.

%%%% Clustering
\subsection*{Clustering}

To assign all images to individual identities, both for automated identity assignment in field monitoring and for evaluating whole-database identity recovery, we first convert the LightGlue matches of each image pair into a pairwise image similarity score and then cluster the images by this score in two steps.

\paragraph{Confidence-filtered similarity score.}
We observed that individuals with many spots tend to yield more keypoints, and that images of many-spotted individuals often produce high keypoint match counts regardless of true identity.
This creates two problems.
First, a global match-count threshold for open-set binary decisions (same vs.\ different) is not meaningful, as the threshold would need to vary with the number of spots on each individual.
Second, in closed-set top-$k$ evaluation, a many-spotted gallery individual can accumulate a higher match count (and therefore a higher similarity score) against a query image than the true match does, if the latter has fewer spots, causing ranking errors. This motivated a similarity score that is robust to the number of spots, i.e., comparable across images of individuals with different numbers of spots.

Denote by $\mathcal{M}(i,j)$ the set of keypoint matches predicted by SuperPoint-LightGlue for image pair $(i,j)$, and let $0 \le p_m \le 1$ denote the confidence score of keypoint match $m$. The confidence-filtered match count at threshold $c$ is
\begin{equation}
    \text{score}_c(i,j) = \left\lvert \{m \in \mathcal{M}(i,j) : p_m \geq c\} \right\rvert
    \label{eq:score_c}
\end{equation}
By default, LightGlue retains only matches with $p_m \geq 0.10$, so $\text{score}_{0.10}(i,j) = \lvert\mathcal{M}(i,j)\rvert$ is the image similarity score used by the experimental pipeline. The optimized pipeline uses $\text{score}_{0.65}$.
Among the alternative similarity scores we considered are the match fraction, $\text{frac}_c(i,j) = \text{score}_c(i,j) / \lvert\mathcal{M}(i,j)\rvert$, and the mean confidence, $\text{mean-confidence}(i,j) = \sum_{m \in \mathcal{M}(i,j)} p_m / \lvert\mathcal{M}(i,j)\rvert$. 

\paragraph{Two-tier clustering.}
Our clustering approach uses two thresholds on the similarity score, $\tau_\text{1}$ and $\tau_\text{2}$ (because $\text{score}_c$ is a count, both are numbers of confident keypoint matches).
In the first (\emph{must-link}) step, any pair of images with high enough similarity, $\text{score}_{c} \geq \tau_\text{1}$, is merged into the same identity via union-find, without requiring agreement from other cluster members.
In the second step, hierarchical clustering starts from the must-link groups and the remaining single images as initial clusters, so must-link groups are never split but can still be merged with each other and with other images; clusters are compared by average linkage, i.e., the mean pairwise $\text{score}_{c}$ across all image pairs between two clusters.
Clusters are merged while the mean $\text{score}_{c}$ between them exceeds $\tau_\text{2}$; the process stops when no remaining pair of clusters exceeds this threshold.

\paragraph{Screening candidate similarity scores.}
We assembled a curated pair dataset from the validation set: all same-identity pairs (3,069 pairs) and an equal number of hard negatives, defined as each query's highest-scoring different-identity candidate according to the experimental pipeline (i.e.\ by match count with the default confidence threshold, $\text{score}_{0.10}$).
We screened 12 candidate similarity scores by average precision (AP; see \emph{Evaluation}), covering confidence statistics, confidence-filtered counts and fractions, keypoint-count normalisation, and geometric verification (Table~\ref{tab:screening}).
The two best-performing score families were carried forward for full-gallery validation: confidence-filtered counts and fractions ($\text{score}_c$ and $\text{frac}_c$) and confidence statistics (represented by $\text{mean-confidence}$); for $\text{frac}_c$, $c$ was fixed at $0.65$, the value that maximised its AP over $c = 0.10$ to $0.95$ (steps of $0.05$), although AP varied by less than $0.001$ between $c = 0.55$ and $0.80$.

\paragraph{Full-gallery validation.}
A curated balanced sample of hard negatives cannot expose failure modes that only appear against a realistically imbalanced gallery. We therefore evaluated the carried-forward candidates, together with default match count as a baseline, on all within-site pairs in the validation set ($\sim$1.09M pairs, of which $0.28\%$ are same-identity pairs).
Candidates were compared by top-1 accuracy and ARI (see \emph{Evaluation}).
Top-1 accuracy was computed by querying each image against all other images from the same site. To estimate its variability, individuals were divided into five disjoint groups, top-1 accuracy was computed with each group's images as queries (the gallery always being the full set), and we report the mean $\pm$ SD over the five groups; no parameter was fitted on these groups.
Because clustering assigns identities jointly across the whole gallery, ARI cannot be meaningfully computed per group, and we report it once over the full validation set.
To compare the candidate scores under the same procedure, each was paired with average linkage clustering, i.e., only the second step of the final clustering procedure, and we report the highest ARI over a fixed grid of $\tau_\text{2}$ values (Table~\ref{tab:scoring}).
The must-link step was omitted because it requires a threshold at which must-link pairs reach 99\% precision (see \emph{Parameter selection}), and no such threshold exists for $\text{frac}_{0.65}$.

\paragraph{Parameter selection.}
For $\text{score}_c$, the confidence threshold $c$ was selected first, by running the two-tier clustering at six candidate thresholds, $c = 0.60$ to $0.85$ in steps of $0.05$,
and taking the value that maximised ARI rather than retrieval accuracy alone.
For each candidate $c$, $\tau_\text{1}$ was set by the precision rule below and $\tau_\text{2}$ by maximising ARI over a coarse grid (steps of 10).

The clustering thresholds were then set for the selected similarity score, $\text{score}_{0.65}$, on the full validation set.
$\tau_\text{1}$ was set to the lowest threshold at which must-link pairs reached at least 99\% precision ($\tau_\text{1} = 55$; precision $99.0\%$, recall $78.3\%$).
$\tau_\text{2}$ was then set to the value that maximised ARI in a fine sweep (steps of 2) with the must-link step applied ($\tau_\text{2} = 29$, ARI $= 0.957$; the coarser step-10 grid used for selecting $c$ peaks at $\tau_2 = 30$, ARI $= 0.956$).
Because these ARI values were maximised on the validation set, they are useful for comparing alternatives, but not for estimating the performance of the optimized pipeline, whose thresholds were fixed and then applied unchanged to the held-out test set and the full dataset.

We chose average linkage over complete linkage because it depends less on the must-link step (see Results).
Single linkage was not considered: at threshold $\tau$ it merges the connected components of pairs with similarity score $\geq \tau$, which is exactly what the must-link step does, so combining the two would amount to a single must-link step at the lower threshold.

%%%%% Evaluation
\subsection*{Evaluation}
\paragraph{Near Eastern fire salamander.}
We evaluated re-identification accuracy on the Near Eastern fire salamander datasets in a closed-set setting. Each image with at least one other image of the same individual in the gallery served in turn as a query, with only the query image itself excluded from the gallery. On the validation set and the full dataset, the gallery comprised all other images from the same site; on the held-out test set, it comprised the other test images from the same site, and in a second evaluation also the remaining images of those sites as distractors.
Same-date images of the same individual were retained in the gallery, as visual inspection confirmed that such pairs can differ in background and posture, and represent realistic field conditions in which the same individual may be encountered multiple times within a single survey night.
Because same-night pairs may be easier, we also report accuracy with same-night images of the query's individual excluded from the gallery (Table~\ref{tab:persite}).

Gallery images were aggregated per identity by their maximum similarity score, and top-$k$ accuracy is the proportion of queries whose true identity is among the $k$ highest-ranked identities (top-1 is the same whether images or identities are ranked).
Ties in similarity score were broken in the query's favour, since both tied candidates would be presented to the observer; on the full dataset, ties decided the top-1 outcome of 3 of the $1{,}855$ queries ($0.16$ percentage points).
We also report mAP@R, the mean over queries of the average precision over the $R$ highest-ranked gallery images, where $R$ is the number of other images of the query's individual in the gallery \citep{musgrave2020metric}; it rewards ranking all images of the true individual near the top, not only the first.
Per-site accuracy is reported in Table~\ref{tab:persite}.

To summarise how well the similarity score separates same- from different-identity pairs across a whole gallery, we use average precision (AP): all within-site pairs are ranked by score, and at each same-identity pair we compute the fraction of pairs ranked at or above it that are also same-identity pairs; AP is the mean of these fractions over all same-identity pairs. We used AP rather than ROC-AUC because, with only about $0.3\%$ of pairs being same-identity pairs, ROC-AUC is dominated by the many different-identity pairs with low scores and is insensitive to false links among the high scores \citep{saito2015precision}.

Top-$k$ accuracy and mAP@R evaluate a ranked list from which an observer confirms each match. Capture--recapture analysis, however, requires a partition of all images into individuals, and assigning this partition automatically removes the observer from the loop. 
We therefore also evaluated the clusters themselves, using the Adjusted Rand Index (ARI; \citealt{hubert1985comparing}), which compares two partitions by counting the image pairs on which they agree, placed together or apart, corrected for chance (ARI $= 1$: perfect agreement; ARI $= 0$: chance level).
Because each pair of images placed together is in effect a recapture link, ARI penalises both splitting errors, which create spurious individuals and lose recaptures, and merging errors, which create false recaptures. ARI is also the evaluation metric of the AnimalCLEF 2026 clustering task \citep{adam2026overview}.
Clustering was performed separately for each site, and ARI was computed once over the combined clusters of all sites in an evaluation set.
Unlike in the top-$k$ evaluation, individuals photographed only once were included.

\paragraph{European fire salamander.}
To evaluate the generalisation of the pipeline beyond the study population and species, we applied it without modification, including all hyper-parameters, to two European fire salamander datasets from the Czech Republic~\citep{adam2025overview} and Germany~\citep{schulte2024performance}.
Top-1 and top-5 accuracy, mAP@R and ARI were computed as for the Near Eastern datasets, with each image that has at least one other image of the same individual serving as a query against all other images of its dataset; the two datasets, collected in different countries, formed separate galleries.
For the German dataset, we additionally computed the two metrics reported by \citet{schulte2024performance}: the recognition rate (RR) and the false rejection rate at rank~1 (FRR$_1$).
RR is the number of recaptures (images other than the first of each individual) correctly identified, divided by the number of recaptures ($63$), where a recapture was counted as correctly identified if its top-ranked candidate was the same individual.
\citet{schulte2024performance} did not state their criterion for a correct identification (e.g., how many top-ranked candidates were inspected); our top-1 criterion is the strictest possible, so the comparison is conservative.
Each recapture was matched only against the images preceding it in the dataset, as in a field study in which each new capture is compared with the existing library; because capture dates were not available, the order of images in the dataset was used as the capture order.
FRR$_1$ is the proportion of query images (images with at least one other image of the same individual) whose top-ranked candidate was not the same individual, i.e.\ $1 -$ top-1 accuracy \citep{bolger2012computer, matthe2017comparison}.
\citet{schulte2024performance} did not state their denominator, but each of their reported values is a whole number of images out of their $377$ images: the failures were divided by all images, including images with no other image of the same individual, which cannot be misidentified.
For comparison with their results, we computed FRR$_1^*$ in the same way, dividing the failed queries by all $376$ images.

% \newpage
%%%%%%%%%%%%%%%%%%%%%%%%%%%%%%%%%%%%%%%%%%%%%%%%
\section*{Results}

\subsection*{Identity audit with the experimental pipeline} 

During the development of our re-identification pipeline, we observed a consistent gap between top-1 and top-2 identity accuracy ($81.7\%$ and $91.3\%$ respectively on the full dataset). We hypothesised that in many of these cases, both top-2 candidates were correct matches split to two different identity labels, such that the pipeline was surfacing labelling errors.
This motivated a systematic audit of the identity labels.

\paragraph{Initial expert review.}
We identified all queries whose labelled identity appeared at rank 2 or 3 in the pipeline ranking and, for each, took every higher-ranked candidate identity. Pairs that arose in both directions were counted once, yielding 51 unique query--candidate identity pairs for expert review; only identities with at least two images can be flagged this way.
For each query, we presented the query image alongside all higher-ranked candidates (rank 1 only if the label appeared at rank 2; ranks 1 and 2 if the label appeared at rank 3), and asked expert ecologists (Rybak, Goldstein) to judge, for each candidate, whether it depicted the same individual as the query.
The experts confirmed 31 of the 51 pairs as depicting the same individual; these corresponded to 26 individuals whose labels were merged, removing 29 redundant labels (splitting errors).
The review also revealed 4 labels that combined images of more than one individual; these were split (merging errors), and in two of them the separated images belonged to an individual that already had a label and were merged into it.
The remaining 20 pairs were not confirmed as matches.
% results: 1,404 id labels 

\paragraph{Stage 1: Intra-identity verification.}
To identify false merges, where a single label was assigned to images of more than one identity, we computed the minimum pairwise similarity score among all image pairs within each identity.
A low minimum similarity indicates that at least two images within the label are dissimilar, suggesting the label may contain more than one identity.
The distribution of minimum similarities showed a small cluster of low-scoring labels separated from the main distribution by a local minimum; we set the threshold at the local minimum of a kernel density estimate of these similarity scores, $T=63$ (Figure~\ref{fig:FigAudit}), flagging $61$ identities for expert review.
Of these, three were confirmed as containing distinct identities and were split to separate identities. % results 1,407 ID labels

\paragraph{Stage 2: Cross-identity verification.}
To identify false splits, where multiple labels were assigned to images of the same identity, we retrieved the top-3 highest-similarity cross-identity matches for each identity.
A human annotator screened all retrieved pairs and judged whether the two labels depicted the same individual; pairs judged to be the same individual were then reviewed by expert ecologists (Rybak, Goldstein) for confirmation.
This stage yielded $316$ identity merges. 
During Stage 2 expert review, three additional intra-identity splits were identified; these had minimum intra-identity similarity scores above the Stage 1 threshold and were therefore not flagged in Stage 1, but were detected incidentally during expert inspection of the label gallery.
The curated dataset comprises $2,440$ images of $1,094$ verified individuals, used for all subsequent development and evaluation.
Nearly all corrections were made in the NTD datasets.
The audit reduced the number of identities from $832$ putative individuals to $553$ verified individuals in NTD\,2022--23, the larger dataset, and from $379$ to $322$ in NTD\,2021--22, whereas in the Haifa datasets it changed only a single label.

Following label correction, the experimental pipeline achieved top-1 and top-2 accuracies of $96.3\%$ and $97.3\%$ on the full dataset, reducing the gap between them from $9.6$ to $1.0$ percentage points.

\subsection*{Pipeline optimization on the validation set}
All scoring and parameter decisions (choice of similarity score, match confidence threshold $c$, keypoint budget $k$ and clustering thresholds $\tau_1$, $\tau_2$) were made using the validation set ($984$ individuals, $2,182$ images, see \emph{Methods}). 
Results in this section are therefore reported on the validation set only.
Because each configuration was selected as the best among several candidates, these values support the decisions but may overestimate performance on held-out individuals.
The optimized pipeline is therefore evaluated on the held-out test set and the full dataset in the next section.

\begin{figure}
    \begin{subfigure}{0.49\linewidth}
        \begin{tikzpicture}
            \node[anchor=south west, inner sep=0] (image) at (0,0)
                {\includegraphics[height=8.5cm,  keepaspectratio]{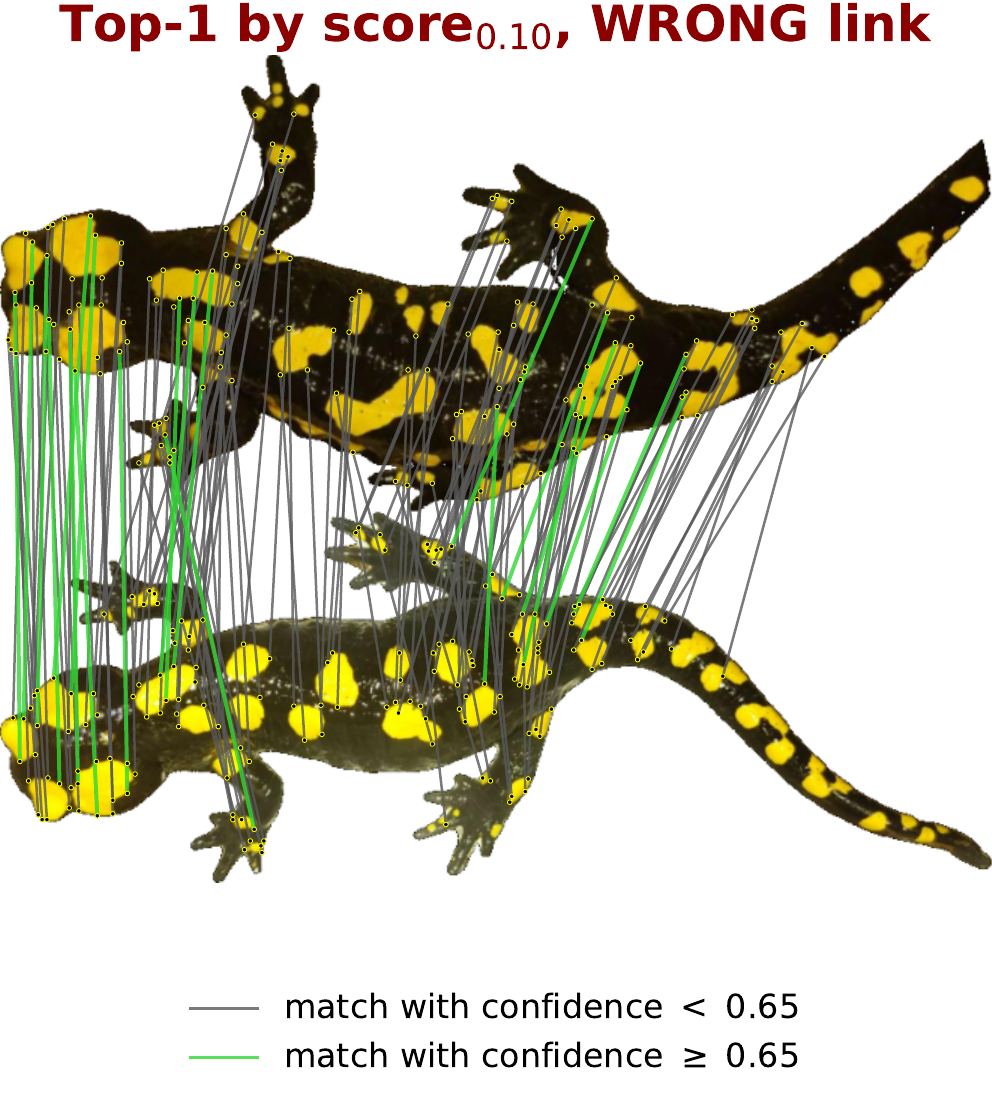}};
            \node[anchor=north west, yshift=-1em, fill=white, fill opacity=0.6,
                  text opacity=1, font=\small] at (image.north west) {(a)};
        \end{tikzpicture}
    \end{subfigure}
    \hfill
    \begin{subfigure}{0.49\linewidth}
    \begin{tikzpicture}
        \node[anchor=north west, inner sep=0] (image) at (0,0)
            {\includegraphics[height=8.5cm, keepaspectratio]{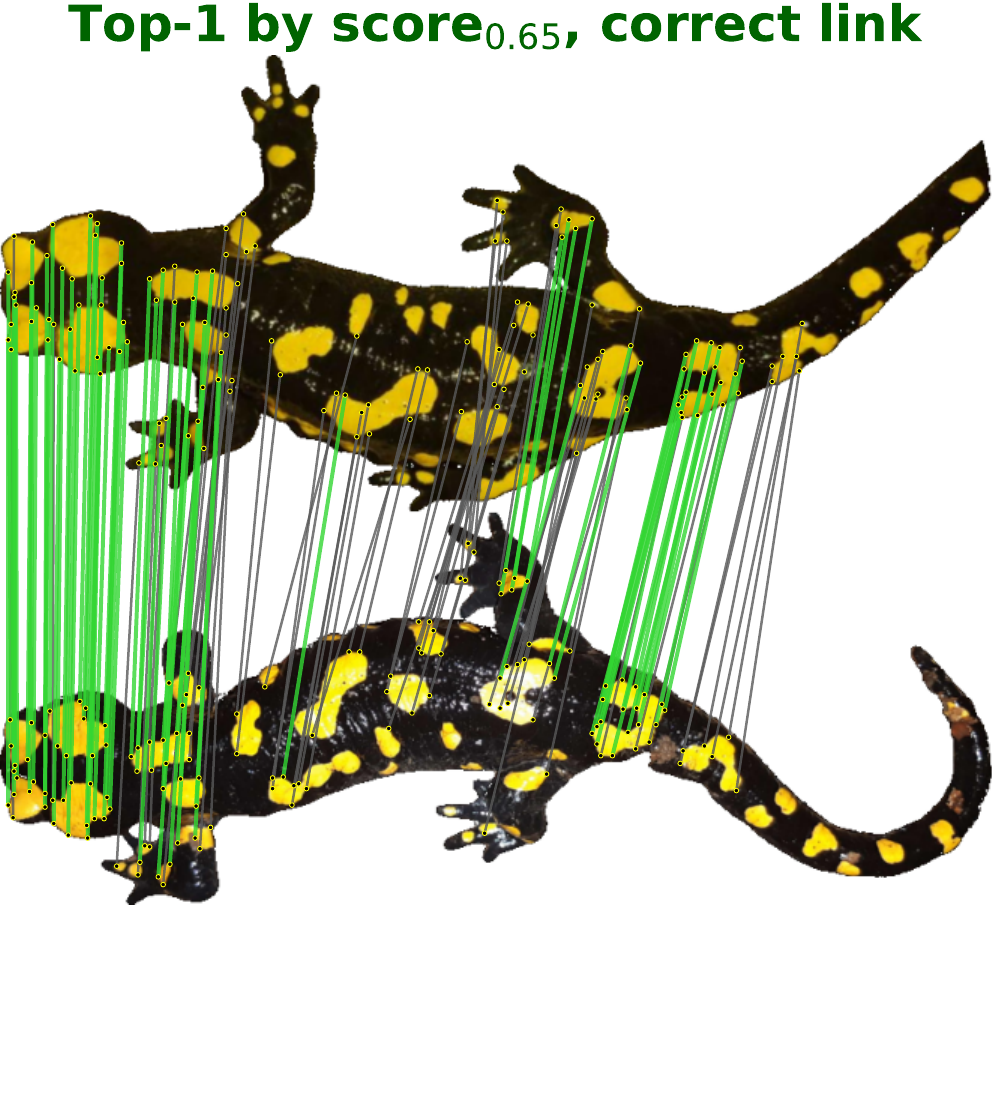}};
        \node[anchor=north west, yshift=-1em, fill=white, fill opacity=0.6,
              text opacity=1, font=\small] at (image.north west) {(b)};
    \end{tikzpicture}
    \end{subfigure}
    \caption{\textbf{Example of the many-spot confound and its correction by confidence filtering.}
    Both panels show the same query (individual 349 from NTD\,2021--22, on top) matched against two candidates from the same dataset.
    Grey lines mark keypoint matches with confidence $\in [0.10, 0.65)$, which contribute to $\text{score}_{0.10}$ but are excluded by $\text{score}_{0.65}$; green lines mark matches with confidence $\geq 0.65$, counted by $\text{score}_{0.65}$. \textbf{(a)} Top-1 candidate by $\text{score}_{0.10}$: a busy false candidate (individual 307, bottom), with $\text{score}_{0.10} = 161$ but $\text{score}_{0.65} = 21$; most keypoint matches are low-confidence.
    \textbf{(b)} True match from a different survey night, ranked second (top-2) by $\text{score}_{0.10} = 141$ but first by $\text{score}_{0.65} = 72$. 
    Using LightGlue's default confidence threshold $c=0.1$ incorrectly ranks (a) above (b); $\text{score}_{0.65}$ correctly recovers the true identity.
    }
    \label{fig:mechanism}

    \vspace{2em}
    % Table kept in the same float as Figure~\ref{fig:mechanism} so they share a page
    {\centering
    \captionof{table}{\textbf{Validation of candidate similarity scores on the validation set.}
    $\text{frac}_{c}$ and $\text{mean-confidence}$ collapse under realistic class imbalance despite competitive performance on the balanced sampled pairs.
    $^*$~Highest ARI over a fixed threshold grid with average linkage alone.}
    \label{tab:scoring}
    \tablestyle
    \begin{tabular}{lccc}
    \tabtop
    \textbf{Similarity score} & \textbf{top-1 (\%)} & \textbf{top-1 std. dev.} & \textbf{ARI$^*$} \\
    \tabhead
    $\text{score}_{0.10}$ (default) & 95.59 & 0.92 & 0.904 \\
    $\text{mean-confidence}$               & 91.39 & 1.97 & 0.879 \\
    $\text{frac}_{0.65}$                    & 88.04 & 2.73 & 0.842 \\
    $\text{score}_{0.65}$ (selected)        & \textbf{97.73} & \textbf{0.52} & \textbf{0.944} \\
    \tabbottom
    \end{tabular}
    \par}
\end{figure}

\paragraph{Confidence filtering reduces the spot-density confound.}
Inspecting the top-ranked candidates of the experimental pipeline, we observed that false candidates were often many-spotted individuals: such individuals yield more keypoints, so pairs of them can accumulate many matches regardless of identity. 
In the example of Figure~\ref{fig:mechanism} (400 keypoints), a busy false candidate collects $161$ matches against the query and outranks the true match ($141$), yet only $21$ of its matches reach confidence $0.65$, compared to $72$ for the true match. 
Confidence filtering can therefore remove precisely the matches that inflate false candidates.
On the validation set, increasing the LightGlue confidence threshold from the default $c=0.1$ to $c=0.65$ increased top-1 accuracy from $95.59\%$ to $97.73\%$, halving the error rate, and reduced its variability across query groups (standard deviation dropped from $0.92$ to $0.52$; Table~\ref{tab:scoring}).

\paragraph{Confidence filtering also improves clustering.}
With average-linkage clustering alone, the increased confidence threshold $(c=0.65)$ raised ARI from $0.904$ (with $c=0.1$) to $0.944$ (Table~\ref{tab:scoring}). Adding the must-link step raised it further to $0.957$, whereas for $c=0.1$ the must-link step slightly lowered ARI ($0.900$).
The must-link step merges any pair above a high similarity score without consulting other images, so it helps only when high similarity scores are reliable evidence of identity.
This is consistent with the spot-density confound: a high similarity score with a low match confidence threshold ($c=0.1$) can also arise between two many-spotted individuals, whereas a high score with a high threshold rarely does. 
Among the six confidence thresholds tested ($c = 0.60$--$0.85$), $c = 0.65$ gave the highest ARI and tied for the best top-1 accuracy, but both varied little across the sweep (ARI $0.94$--$0.96$, top-1 $97.2$--$97.7\%$), so the results are insensitive to the exact choice of threshold $c$.

\paragraph{Clustering is stable around the selected thresholds.}
With the must-link step applied, ARI remained between $0.951$ and $0.957$ for $23 \le \tau_\text{2} \le 35$ (steps of 2), so the choice of $\tau_\text{2} = 29$ is insensitive to small perturbations. Complete linkage reached an ARI of $0.956$ once the must-link step was applied, equivalent to that of average linkage, but depended on it far more: without must-link, its ARI was only $0.848$ (a gain of $+0.108$ from must-link), compared with $0.944$ for average linkage ($+0.013$). Average linkage is therefore the more robust choice on the validation set. 

\paragraph{Similarity scores that ignore the number of keypoint matches collapse under realistic class imbalance.}
On the curated pair sample, several alternative similarity scores reached slightly higher average precision than the confidence-filtered count (AP $= 0.954$--$0.957$ vs.\ $0.951$; Table~\ref{tab:screening}).
On the full validation set, where only $0.28\%$ of pairs are same-identity pairs, the two alternative similarity scores carried forward both fell below $\text{score}_{0.10}$ in top-1 accuracy ($\text{mean-confidence}$: $91.39\%$; $\text{frac}_{0.65}$: $88.04\%$; Table~\ref{tab:scoring}).
Both scores ignore how many matches a pair has, so a pair with a few confident matches can rank above a same-identity pair with many matches at moderate confidence.
Such pairs are rare among the hardest negatives of the curated sample, one per query, but common among the $\sim$1.09M pairs of the full gallery. Screening on a curated sample is cheap and highlights the hard negatives, but it can be misleading unless it is followed by validation against a realistically imbalanced gallery.

% === Evaluation - test set & full dataset
\subsection*{Evaluation of the optimized pipeline}

With all choices fixed, the optimized pipeline was evaluated on two partitions (Table~\ref{tab:results}). The held-out test set ($110$ individuals, $258$ images; AH: $193$, HFA: $65$) was used once, to test whether these choices generalise to individuals that were not used for tuning. 
The full dataset ($1{,}094$ individuals, $2{,}440$ images) reflects the realistic setting, in which each query is matched against all other images from the same site, and allows a direct comparison with the experimental pipeline. Because the full dataset includes the validation set, its results may be optimistic.

\paragraph{The optimized pipeline maintained high accuracy on held-out individuals.}
In the held-out test set, $206$ of the $258$ images have at least one other image of the same individual within the test set; these served as queries against the other test images.
The optimized pipeline reached $95.15\%$ top-1 and $95.63\%$ top-5 accuracy (mAP@R $= 0.884$). Adding the other $252$ AH and HFA images to the gallery as distractors left top-1 and top-5 unchanged and lowered mAP@R to $0.881$.

On the same images and keypoints, using the match count at the default confidence threshold ($\text{score}_{0.10}$) as the similarity score yielded $96.12\%$ top-1 and $97.09\%$ top-5 accuracy (mAP@R $= 0.877$), two and three more correct queries out of $206$ than $\text{score}_{0.65}$.
However, mAP@R, which accounts for the ranks of all true references, favoured $\text{score}_{0.65}$.
The test set contains only AH and HFA individuals, whose per-site galleries are small ($193$ and $65$ images).
The spot-density confound requires a query to compete with many many-spotted candidates, which is more likely in large galleries such as NTD\,2022--23 ($1{,}340$ images), so the benefit of confidence filtering may be concentrated at the NTD sites, which the test set does not include.

\begin{table}[h]
\centering 
\caption{\textbf{Performance of the optimized pipeline.} Held-out test: $206$ queries (test images with at least one other image of the same individual in the test set), each matched against the other test images. Full dataset: $2{,}440$ images of $1{,}094$ individuals, including the validation set; $1{,}855$ queries, each matched against all other images from the same site. Time: keypoint extraction and matching on a single GPU, excluding preprocessing. ARI: two-tier clustering; thresholds were fixed on the validation set for the optimized pipeline and calibrated on the experimental pipeline's own similarity scores for the experimental pipeline.}
\label{tab:results}
\tablestyle
\setlength{\tabcolsep}{4pt}
\begin{tabular}{llrrcccc}
\tabtop
\textbf{Evaluation set} & \textbf{Pipeline} & \textbf{Queries} & \textbf{Time (s)} & \textbf{top-1 (\%)} & \textbf{top-5 (\%)} & \textbf{mAP@R} & \textbf{ARI} \\
\tabhead
\grp{Held-out test} & Optimized & 206 & - & 95.15 & 95.63 & 0.884 & 0.958 \\
\tabsep
\grp{Full dataset} & Experimental & 1,855 & \textbf{34,509} & 96.33 & \textbf{98.22} & 0.868 & 0.892 \\
 & Optimized & 1,855 & 35,108 &\textbf{97.47} & 98.11 & \textbf{0.908} & \textbf{0.958} \\
\tabbottom
\end{tabular}
\end{table}

\begin{figure}
    \centering 
    \includegraphics[width=0.5\linewidth]{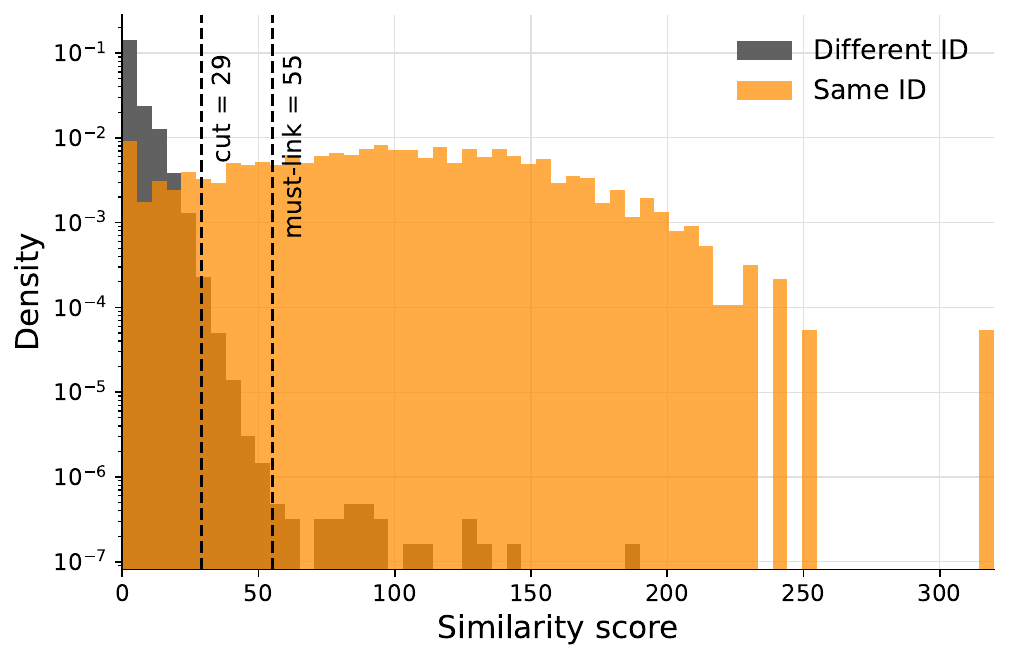}
    \caption{\textbf{Same-identity pairs have higher similarity scores than different-identity pairs.}
        Distribution of pairwise similarity scores ($\mathrm{score}_{0.65}$) for the optimized pipeline on the full dataset (2,440 images of 1,094 individuals). Each within-site image pair contributes one score; pairs are never formed across sites. Same-identity pairs (orange, $n=3{,}453$) and different-identity pairs (grey, $n=1{,}146{,}843$) overlap at low scores but not at high scores. Density is plotted on a logarithmic axis. Dashed lines mark the two clustering thresholds, the average-linkage cut ($29$) and the must-link threshold ($55$), both fixed in advance on the validation set. The must-link threshold admits $77.4\%$ of same-identity pairs while excluding all but $0.0021\%$ of different-identity pairs.}

    \label{fig:histogram}
\end{figure}

\begin{figure}[!ht]
    \centering
    \includegraphics[width=0.5\linewidth]{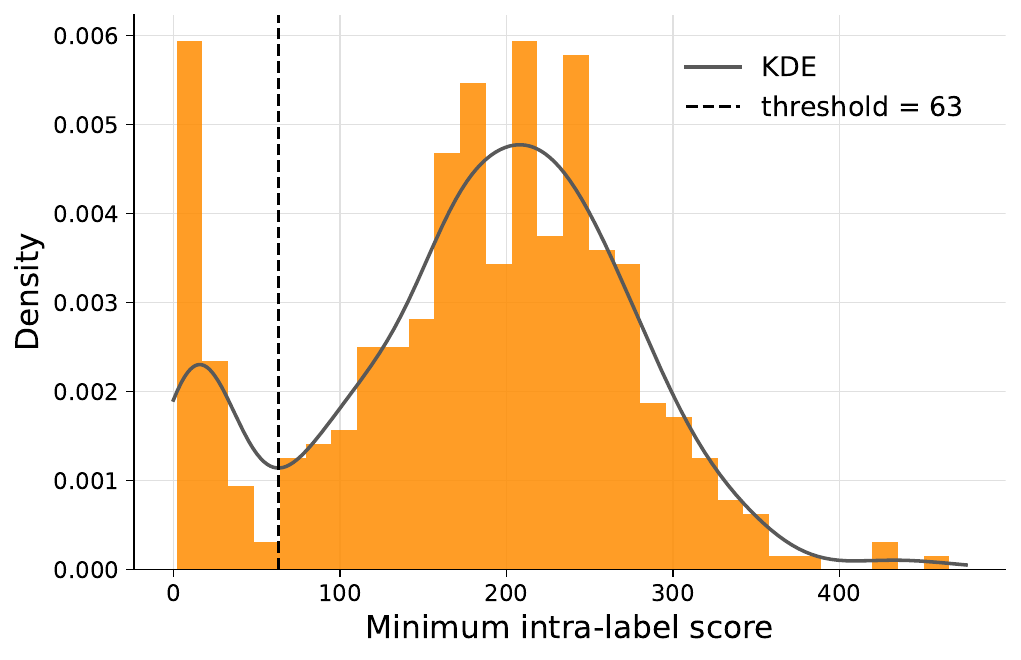}
    \caption{\textbf{
        Distribution of minimum pairwise image similarity scores across all identities with two or more images ($n=414$).} 
        The solid line shows a KDE of the score distribution, revealing a small cluster of low-scoring identities separated from the main distribution by a local minimum, indicative of possible label errors.
        The dashed line marks the expert review threshold, set at the local minimum of the KDE ($T=63$).}
    \label{fig:FigAudit}
\end{figure}

\paragraph{Most same-identity pairs can be linked automatically with high precision.}
Of the $1{,}150{,}296$ within-site pairs in the full dataset, only $3{,}453$ ($0.3\%$) are same-identity pairs (Figure~\ref{fig:histogram}). 
Ranked by $\text{score}_{0.65}$, these pairs reached AP $= 0.890$ (AP $= 1$ means every same-identity pair outranks every different-identity pair); with $\text{score}_{0.10}$, AP was $0.829$, because more different-identity pairs outranked same-identity pairs.
The two pairwise similarity score distributions do not overlap at high similarity scores:
at the must-link threshold ($\tau_\text{1} = 55$), $77.4\%$ of same-identity pairs were linked, whereas only $0.0021\%$ of different-identity pairs were linked (about $24$ pairs), a precision of $99.1\%$.
The remaining same-identity pairs score within the range of different-identity pairs; they can be grouped only by links to other images of the same individual, which is the role of the average-linkage step.

\paragraph{Automatic clustering recovers individual identities.}
The two-tier clustering combines both kinds of evidence: it first merges the high-precision must-link pairs and then groups the remaining images by average linkage.
With thresholds fixed on the validation set, where they were tuned to maximise ARI ($0.957$), it achieved ARI $= 0.958$ on both the held-out test set and the full dataset, compared to $0.892$ for the experimental pipeline on the full dataset.
Clustering and retrieval nonetheless answer different questions: on the validation set, assigning each query the majority identity of its cluster gave $96.0\%$ top-1 accuracy, below the $97.73\%$ of similarity-based ranking.
A cluster-based decision depends on all pairs in the cluster, whereas a ranking-based decision depends only on the highest-ranking reference.
We therefore use ranking to identify single queries and clustering to reconstruct the identities of the whole database, as required for downstream capture--recapture analysis.

\subsection*{Generalization to the European fire salamander}

Applied without any modification to the two \eurosalamander\ datasets, the optimized pipeline achieved high top-$k$ accuracy, mAP@R and ARI (Table~\ref{tab:generalization}), despite the more variable markings of this related species (Figure~\ref{fig:Fig1}c,d).
On the Czech dataset, the pipeline achieved top-1 $95.07\%$, top-5 $97.09\%$ and mAP@R $= 0.918$.

\paragraph{The pipeline outperformed five established photo-identification tools.}
On the German dataset ($376$ images, $63$ recaptures), the pipeline achieved top-1 accuracy of $91.7\%$ and RR $= 93.7\%$, outperforming all five photo-identification tools tested by \citet{schulte2024performance}, whose RR ranged from $37.9\%$ to $77.3\%$. Unlike four of these tools, which required manual image preprocessing (cropping, placing reference points or coding pattern characters), our pipeline preprocesses images automatically.
Its recognition rate was close to that of visual inspection by two observers ($93.9\%$).
Computed as in their study (FRR$_1^*$; see Methods), our false rejection rate at rank~1 was $2.4\%$, compared with $9.8$--$15.1\%$ for the five tools.
\citet{schulte2024performance} counted $66$ recaptures, whereas we evaluated $63$ after removing one duplicate image and three double-counted recaptures.
This difference does not affect the comparison: counting the three removed recaptures as misses would give RR $= 89.4\%$, still well above the best tool from \citet{schulte2024performance}. % 59/66=89.4% 

\begin{table}[h]
\centering
\caption{\textbf{Generalization to the European fire salamander datasets.} 
RR and FRR$_1^*$ are reported only for the German dataset. FRR$_1^*$ was computed as by \citet{schulte2024performance}, dividing rank-1 failures by all images, for comparison with their results; under the common definition ($1 -$ top-1 accuracy), FRR$_1 = 8.3\%$.}
\label{tab:generalization}
\tablestyle
\begin{tabular}{lcccccc}
\tabtop
\textbf{Dataset} & \textbf{top-1 (\%)} & \textbf{top-5 (\%)} & \textbf{mAP@R} & \textbf{ARI} & \textbf{RR (\%)} & \textbf{FRR$_1^*$ (\%)} \\
\tabhead
Czech dataset & 95.07 & 97.09 & 0.918 & 0.885 & -- & -- \\
German dataset & 91.74& 95.41 & 0.861 & 0.768 & 93.65& 2.4 \\ 
\tabbottom
\end{tabular}
\end{table}

\paragraph{Ranking transferred better than clustering thresholds.}

Comparing the held-out Near Eastern fire salamander dataset and the two European fire salamander datasets, we find that retrieval metrics (top-$k$ accuracy, mAP@R) remain broadly comparable across datasets, whereas our clustering metric (ARI) is lower on the Czech dataset (0.885) and substantially lower on the German dataset (0.768) relative to Near Eastern salamander (0.958), see \Cref{tab:generalization,tab:results}.
Accuracy and mAP@R depend on similarity score rankings, whereas ARI also depends on the absolute thresholds ($\tau_1$, $\tau_2$) tuned on our validation set. Differences in similarity distributions across datasets may therefore explain the larger decline in ARI, suggesting that recalibrating these thresholds on a small labelled sample could improve clustering performance.

% \newpage
%%%%%%%%%%%%%%%%%%%%%%%%%%%%%%%%%%%%%%%%%%%%%%%%

\section*{Discussion}
% --- 1. Summary of the main results: the method 
Our pipeline re-identified Near Eastern fire salamanders accurately across four datasets and four sites, with $1{,}094$ individuals.
We only adapted the preprocessing (segmentation, orientation and spot extraction), the similarity score (confidence threshold $c$) and the clustering (thresholds $\tau_1$ and $\tau_2$) to the Near Eastern fire salamander, tuning them on our Israeli validation set only, and made no adaptations for the European fire salamander. The local-feature matching model was used without adaptation to animal re-identification, and its original training used generic internet photographs for homography pretraining and landmark photographs for subsequent finetuning.
Our pipeline ranked the correct individual first for $95.2\%$ of held-out queries ($97.5\%$ on the full dataset) and clustered identities with ARI $= 0.958$ on both. Two choices contributed most: filtering low-confidence keypoint matches, which reduced a confound between spot density and match count and improved clustering and mAP@R, although on the small held-out galleries the unfiltered match count was slightly more accurate at rank~1 (two of $206$ queries); and a two-tier clustering that turns pairwise similarity scores into identities for the whole dataset.
Without any modification, our pipeline re-identified held-out European fire salamanders in datasets from Czechia and Germany. In the latter, our pipeline substantially outperformed five established photo-identification tools.

% --- 2. Summary of the results: practical value for the database 
The pipeline also proved useful for auditing the datasets: many of its apparent errors were mistakes in the labels (human annotations), even though the matcher had no species-specific training. The audit reduced the number of identities by almost a quarter, and nearly all corrections were splitting errors, as in photo-identification studies of other species \citep{stevick2001errors, johansson2020identification, jordan2025supplementing}, and nearly all were made at Tel Dan Nature Reserve, where hundreds of individuals were recorded in each season. In a critically endangered skink, software-assisted matching likewise showed that experienced observers had registered 42 recaptured individuals as new, with no false merges, errors the authors attributed to the difficulty of comparing images against a large catalogue \citep{jordan2025supplementing}. Similarly, human observers misclassified 12.5\% of queries in a study of captive snow leopards (\snowleopard; \citealt{johansson2020identification}). Because splitting errors add spurious individuals and remove recaptures, they bias capture--recapture estimates of population size upward \citep{stevick2001errors, johansson2020identification, kodi2024ghostbusting}. Thus, photo-identification databases should be validated manually before being used for population monitoring, or capture--recapture methods should account for these `ghost' individuals. For example, identities with a single capture (image) can be discarded, thereby reducing bias but increasing variance of the estimated population size \citep{morrison_estimating_2011}.

% --- 3. Comparison with previous methods 
Our results can be placed in the context of three generations of re-identification methods.
The first generation, which includes the programs used in previous studies of salamanders and newts, relies on hand-crafted features: SIFT keypoints in Wild-ID and I3S \citep{bolger2012computer}, HotSpotter in Amphibian \& Reptile Wildbook \citep{crall2013hotspotter}, the number of matching pixels between manually outlined and straightened patterns in AmphIdent \citep{drechsler2015genetic}, and manually coded pattern characters in ManderMatcher \citep{speybroeck2017pattern}.
These programs can be highly accurate when photographs are standardized: in great crested newts photographed flat in a transparent box, AmphIdent missed only $2\%$ of recaptures \citep{drechsler2015genetic}.
But all except Wildbook require manual preprocessing, and their accuracy falls with less standardized photographs \citep{matthe2017comparison}: two of them recognised nearly all recaptures of European fire salamander larvae photographed in the laboratory, but the five recognised only $38$--$77\%$ of recaptures of adults photographed in the field, whose bodies were often bent or stretched \citep{schulte2024performance}.
On the same adult images, our pipeline recognised $93.7\%$ of recaptures without manual preprocessing, close to visual inspection by two observers ($93.9\%$).
The second generation represents the whole image by a single vector computed by a deep network trained on the target species.
A convolutional network fine-tuned on head spots identified $11$ captive Japanese giant salamanders (\japsalamander) photographed on two consecutive days \citep{takaya2023individual}, and similar networks were trained for yellow-spotted mountain newts \citep{rahmdel2026deep}.
Such classifiers recognise only the individuals they were trained on, so each new individual requires labeled images and retraining.
Embedding models pretrained on many species \citep{cermak2024wildlifedatasets, otarashvili2024miewid} avoid retraining but transfer poorly to amphibians: for the Hula painted frog, the best such model ranked the correct individual first for $10.5\%$ of queries, and for $62.1\%$ after fine-tuning on labeled images of the species \citep{yesharim2026latonia}; because the two species present a similar problem, a pattern of spots on smooth skin, we did not repeat this comparison here.
The third generation, to which our pipeline belongs, learns keypoints, descriptors and matches from generic photographs \citep{zhao2023aliked, lindenberger2023lightglue} and applies them to animals without training on the target species \citep{nepovinnykh2024species, yesharim2026latonia}, which addresses the call for one tool applicable across species rather than species-specific programs \citep{drechsler2015genetic}; each match comes with a confidence, on which our scoring is built.
The AnimalCLEF 2025 challenge, which included \eurosalamander, points the same way: nearly all top-ranked systems combined global embeddings with local-feature matching, and the winning system used global embeddings only to rank candidates, which it then re-ranked by local matches \citep{adam2025overview}.

% --- 4. Comparison with the Hula painted frog and other amphibians 
The same principle identified Hula painted frogs (\latonia) from their ventral patterns with $99.8\%$ top-1 accuracy \citep{yesharim2026latonia}. The salamander data were harder: galleries held up to $553$ individuals per site, compared to $191$ in the frog dataset, and individuals were photographed less often ($2.2$ vs.\ $6.5$ images per individual). The different body plan required adapting the preprocessing, in particular extracting the yellow spot pattern before keypoint detection (Figure~\ref{fig:preprocessing}c,f), which restricts the keypoint detector to the identifying pattern rather than skin texture, reflections or mud.

% --- 6. Comparison with the European fire salamander 
We tested transfer of our pipeline from the Near Eastern to the European fire salamander, with different marking pattern ranging from discrete spots to elongated stripes.
Identity ranking transferred well: top-1 accuracy was $95.1\%$ and $91.7\%$.
Clustering transferred less well: ARI fell from $0.958$ to $0.885$ and $0.768$ on the Czech and German datasets. The ranking depends only on the ranking of scores, whereas the clustering thresholds were set on \eastsalamander and may not separate same- and different-identity pairs equally well for another pattern type. Thus, before clustering is used to build capture histories for a new species, its thresholds may need to be validated on a small labeled images dataset of the target species.

% --- 7. Limits for other species 
The approach works best when the image given to the keypoint detector is dense in identifying information. In fire salamanders, isolating the yellow spot pattern (Figure~\ref{fig:preprocessing}c,f) yields such an image: most of what remains is the identifier itself. We expect the approach to be less reliable for species whose identifying marks are sparse, for example when individuals differ only in the position of a few small spots. Isolating such marks leaves too little to match, whereas keeping the whole individual leaves an image dominated by anatomical features shared by all individuals, such as the body outline, eyes and limbs; in both cases, same- and different-identity pairs would separate poorly. The same applies when marks are often hidden by posture or occlusion, or when the database mixes viewpoints, such as close-ups of the head alongside dorsal views, a difficulty we encountered in the AnimalCLEF 2026 fire salamander data \citep{yesharim2026animalclef}.

% --- 8. Limitations of this study 
Some limitations should be noted. The audit was pipeline-assisted: experts confirmed every change, but errors that the pipeline could not detect would remain, so accuracy measured against the verified labels may be slightly optimistic. The held-out test set came only from the AH and HFA sites, whose galleries are small, and contains no Tel Dan individuals. Same-night images were retained in the gallery, whereas the Hula painted frog study discarded same-date pairs \citep{yesharim2026latonia}; this reflects field use but may make some matches easier. 

% --- 9. Closing 
Zero-shot local-feature matching thus offers amphibian monitoring programs an identification tool that needs no training data, transfers between species, and scales to databases of more than a thousand individuals. In our data, the limiting step was not the photographs, which sufficed for experts to confirm every correction, but matching them by eye as the database grew. We therefore recommend checking photo-identification databases by automated clustering followed by expert review before they are used for capture--recapture analysis. Applying the pipeline to link individuals across seasons would extend this work to estimates of survival and population trends.

%%%%%%%%%%%%%%%%%%%%%%%%%%%%%%%%%%%%
% Acknowledgements
{\small
\section*{Acknowledgements}
We thank Oren Kolodny and Eli Geffen for discussions and comments and Sagi Potash for help with auditing.
We thank all volunteers who participated in the salamander project in Haifa.
We thank Laura Schulte and Barbara Caspers for sharing the identity labels of the German dataset.
We thank Luk\'a\v{s} Adam and Luk\'a\v{s} Picek for sharing the identity labels of the AnimalCLEF salamander test set.
We thank Eitan Nissim, manager of Tel Dan Nature Reserve, and the reserve's rangers for conducting surveys and photographing the salamanders.

This work was supported in part by Safra Center for Bioinformatics and the Center for AI \& Data Science, Tel Aviv University (YR), and by Minerva Center for the Study of Population Fragmentation.

\section*{Declarations}
The authors declare no conflict of interest.

ChatGPT and Claude were used to write source code and text, under supervision, validation, and editing by the first and last author.
}

\bibliographystyle{agsm}
\bibliography{references}

\newpage
\beginsupplement
\section*{Supplementary Material}

\begin{table}[!ht]
\centering
\caption{%
\textbf{Average precision (AP) and ROC-AUC for 12 candidate similarity scores on the
curated pair sample from the validation set (no held-out test images)}: all same-identity pairs
($n{=}3{,}069$) balanced against an equal number of hardest negatives
(each query's highest-scoring different-identity candidate under the
experimental pipeline). Confidence-filtered counts and fractions are shown at a fixed screening cutoff, $c = 0.7$.  $n_a,n_b$ is the number of SuperPoint keypoints detected in images $a,b$.
AP is used for consistency with every other screening stage in this
work; AUC is reported as a secondary reference, since this sample is
balanced by construction and so is not itself subject to the class
imbalance AP was adopted to address. Confidence-based scores
(rows 2--8) are consistently strongest; keypoint-count normalization
and geometric verification (rows 9--12), both designed specifically to
counter the busy-spot confound, score no better than the raw baseline
(AP $0.786$--$0.885$ vs.\ $0.882$).
Boldface marks the three-way tie for best AP (0.957).
}
\label{tab:screening}
\tablestyle
\begin{tabular}{@{}lp{6cm}cc@{}}
\tabtop
\textbf{Similarity score} & \textbf{Definition} & \textbf{AP} & \textbf{AUC} \\
\tabhead
\texttt{num\_matches} (raw match count)
  & $\lvert\mathcal{M}(i,j)\rvert$ — all matches, LightGlue's own
    confidence floor ($\geq 0.10$), the experimental pipeline's score
  & 0.882 & 0.822 \\
\texttt{mean\_confidence}
  & mean of $p_m$ over all matches
  & \textbf{0.957} & 0.936 \\
\texttt{median\_confidence}
  & median of $p_m$ over all matches
  & 0.954 & 0.932 \\
\texttt{p70\_confidence}
  & 70th-percentile $p_m$
  & 0.956 & 0.935 \\
\texttt{p90\_confidence}
  & 90th-percentile $p_m$
  & 0.955 & 0.936 \\
\texttt{mean\_top50pct\_confidence}
  & mean $p_m$ over the more-confident half of matches (those $\geq$
    the pair's own median)
  & \textbf{0.957} & 0.937 \\
\texttt{count\_ge\_0.7}
  & $\lvert\{m : p_m \geq 0.7\}\rvert$ — \texttt{count\_c} at one
    fixed cutoff
  & 0.951 & 0.926 \\
\texttt{frac\_ge\_0.7}
  & \texttt{count\_ge\_0.7} $\div$ \texttt{num\_matches}
  & \textbf{0.957} & 0.938 \\
\texttt{match\_ratio\_min}
  & \texttt{num\_matches} $\div$ $\min(n_a, n_b)$
  & 0.872 & 0.817 \\
\texttt{match\_ratio\_mean}
  & \texttt{num\_matches} $\div$ $\mathrm{mean}(n_a, n_b)$
  & 0.879 & 0.820 \\
\texttt{geo\_inlier\_ratio}
  & RANSAC affine-fit inlier count $\div$ \texttt{num\_matches} —
    spatial (not confidence) consistency
  & 0.786 & 0.780 \\
\texttt{geo\_inlier\_count}
  & raw RANSAC affine-fit inlier count
  & 0.885 & 0.854 \\
\tabbottom
\end{tabular}
\end{table}

\begin{table}[!ht]
\centering
\caption{
  \textbf{Identification accuracy of the optimized pipeline per site on the full dataset.} \emph{All}: the gallery is every other image from the same site. \emph{Same date excluded}: images of the query's own individual taken on the same date are also removed from the gallery; images of other individuals from that date remain. $n_\text{eval}$: query images with at least one other image of the same individual remaining in the gallery.}
\label{tab:persite}
\tablestyle
\begin{tabular}{lrr rrrr rrrr}
\tabtop
 & & & \multicolumn{4}{c}{\textbf{All}} & \multicolumn{4}{c}{\textbf{Same date excluded}} \\
\cmidrule(lr){4-7}\cmidrule(lr){8-11}
\textbf{Site} & \textbf{$n_\text{img}$} & \textbf{$n_\text{ids}$} & \textbf{$n_\text{eval}$} & \textbf{top-1} & \textbf{top-5} & \textbf{mAP@R} & \textbf{$n_\text{eval}$} & \textbf{top-1} & \textbf{top-5} & \textbf{mAP@R} \\
\tabhead
AH            &  376 &  182 &  279 & 94.62 & 94.98 & 0.865 &  279 & 94.62 & 94.98 & 0.865 \\
HFA           &  134 &   37 &  123 & 99.19 & 99.19 & 0.961 &  121 & 99.17 & 99.17 & 0.960 \\
NTD\,2021--22 &  590 &  322 &  411 & 97.81 & 98.30 & 0.938 &  403 & 97.77 & 98.26 & 0.937 \\
NTD\,2022--23 & 1340 &  553 & 1042 & 97.89 & 98.75 & 0.902 & 1034 & 97.58 & 98.55 & 0.901 \\
\tabsep
\textbf{Overall} & \textbf{2440} & \textbf{1094} & \textbf{1855} & \textbf{97.47} & \textbf{98.11} & \textbf{0.908} & \textbf{1837} & \textbf{97.28} & \textbf{97.99} & \textbf{0.907} \\
\tabbottom
\end{tabular}
\end{table}

\end{document}